\pdfoutput=1
\documentclass[11pt]{article}

\usepackage[]{ACL2023}

\usepackage{times}
\usepackage{latexsym}

\usepackage[T1]{fontenc}
\usepackage[utf8]{inputenc}

\usepackage{microtype}

\usepackage{inconsolata}

\usepackage{multicol}
\usepackage{booktabs}
\usepackage{makecell}
\usepackage{amsmath, amssymb}
\usepackage{multirow}
\usepackage{mathtools}
\usepackage{xcolor}
\usepackage{enumitem}
\usepackage{float}
\usepackage{graphicx}
\usepackage{upgreek}
\usepackage{seqsplit}
\usepackage{color,soul}
\usepackage{arydshln}
\usepackage{placeins}
\usepackage{pifont}
\usepackage{amssymb}
\newcommand{\cmark}{\ding{51}}%
\newcommand{\xmark}{\ding{55}}%

\usepackage{caption}
\usepackage{subcaption}
\definecolor{realblue}{RGB}{0,0,255}

\title{Phrase Retrieval for Open-Domain Conversational Question Answering with Conversational Dependency Modeling via Contrastive Learning}

\author{Soyeong Jeong$^1$
        \quad Jinheon Baek$^2$
        \quad Sung Ju Hwang$^{1, 2}$
        \quad Jong C. Park$^1$\thanks{\hspace{0.2cm}Corresponding author} \\
        School of Computing$^1$ \quad Graduate School of AI$^2$ \\
        Korea Advanced Institute of Science and Technology$^1$$^,$$^2$\\
       \texttt{\{starsuzi,jinheon.baek,sjhwang82,jongpark\}@kaist.ac.kr}}

\begin{document}
\maketitle
\begin{abstract}

Open-Domain Conversational Question Answering (ODConvQA) aims at answering questions through a multi-turn conversation based on a retriever-reader pipeline, which retrieves passages and then predicts answers with them. However, such a pipeline approach not only makes the reader vulnerable to the errors propagated from the retriever, but also demands additional effort to develop both the retriever and the reader, which further makes it slower since they are not runnable in parallel. In this work, we propose a method to directly predict answers with a phrase retrieval scheme for a sequence of words, reducing the conventional two distinct subtasks into a single one. Also, for the first time, we study its capability for ODConvQA tasks. However, simply adopting it is largely problematic, due to the dependencies between previous and current turns in a conversation. To address this problem, we further introduce a novel contrastive learning strategy, making sure to reflect previous turns when retrieving the phrase for the current context, by maximizing representational similarities of consecutive turns in a conversation while minimizing irrelevant conversational contexts. We validate our model on two ODConvQA datasets, whose experimental results show that it substantially outperforms the relevant baselines with the retriever-reader. Code is available at: \url{https://github.com/starsuzi/PRO-ConvQA}.

\end{abstract}

\section{Introduction}
\begin{figure}[t!]
\begin{center}
\includegraphics[width=0.495\textwidth]{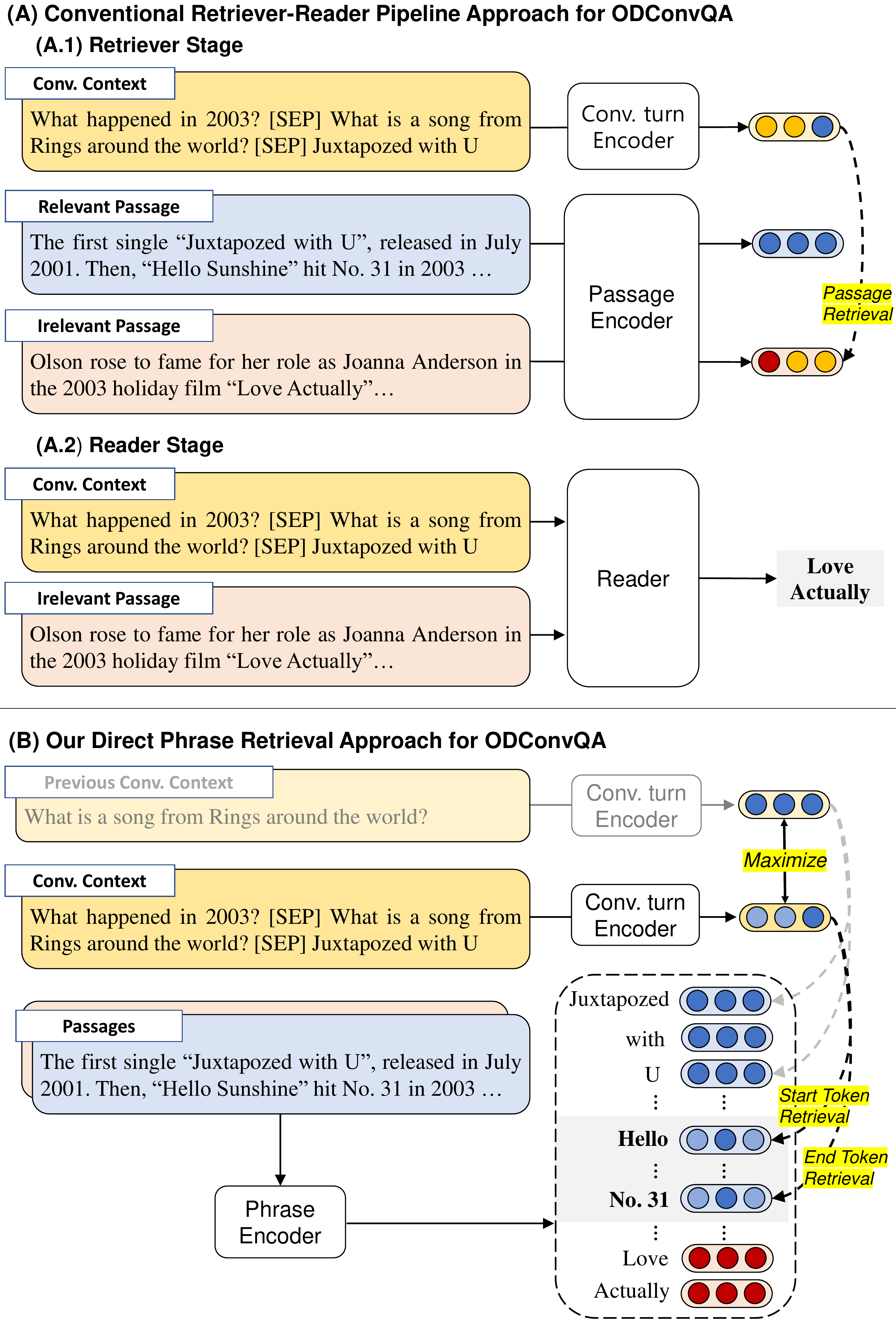}
\end{center}
\vspace{-0.13in}
\caption{
\small (A) Conventional retriever-reader pipeline approach, which first retrieves a relevant passage to a current conversational (i.e., Conv.) context, and then predicts an answer based on the passage. (B) Our direct phrase retrieval approach that predicts start and end tokens of the answer phrase based on their representational similarities to the current Conv. context. To reflect the previous history when retrieving the phrase, we maximize representations of two consecutive conversations.
}
\label{fig:0_concept}
\vspace{-0.2in}
\end{figure}

Conversational Question Answering (ConvQA) is the task of answering a sequence of questions that are posed during information-seeking conversations with users~\cite{DBLP:conf/emnlp/ChoiHIYYCLZ18,DBLP:journals/tacl/ReddyCM19, DBLP:journals/kais/ZaibZSMZ22}. This task has recently gained much attention since it is similar to how humans seek and follow the information that they want to find. To solve this problem, earlier ConvQA work proposes to predict answers based on both the current question and the previous conversational histories, as well as the passage that is relevant to the ongoing conversation~\cite{DBLP:conf/cikm/QuYQZCCI19, DBLP:conf/iclr/HuangCY19, DBLP:conf/acl/KimKPK20, DBLP:conf/acl/LiGGC22}. However, this approach is highly suboptimal and might not be applicable to real-world scenarios, since it assumes that the gold passage, containing answers for the current question, is given to the ConvQA system; meanwhile, the gold passage is usually not available during the real conversation.

To address this limitation, some recent work~\cite{DBLP:conf/sigir/Qu0CQCI20,DBLP:conf/naacl/AnanthaVTLPC21,  DBLP:journals/tois/LiLN22, DBLP:journals/tacl/AdlakhaDSVR22, DBLP:conf/ijcnlp/FangHHC22} proposes to extend the existing ConvQA task to an open-domain question answering setting with an assumption that the conversation-related passages are not given in advance; therefore, it is additionally required to access and utilize the query-relevant passages in a large corpus, for example, Wikipedia. Under this open-domain setting, most existing Open-Domain ConvQA (ODConvQA) work relies on the retriever-reader pipeline, where they first retrieve the passages, which are relevant to both the current question and conversational context, from a large corpus, and then predict answers based on information in the retrieved passages. This retriever-reader pipeline approach is illustrated in Figure~\ref{fig:0_concept}.

However, despite their huge successes, such a pipeline approach consisting of two sub-modules has a few major drawbacks. First, since the reader is decomposed from the retriever, it is difficult to train the retriever-reader pipeline in an end-to-end manner, which results in an additional effort to develop both the retriever and the reader independently. Second, the error can be accumulated from the retriever to the reader, since the failure in finding the relevant passages for the current question negatively affects the reader in predicting answers, which is illustrated in Figure~\ref{fig:0_concept}. Third, while the latency is an important factor when conversing with humans in the real-world scenarios, the retriever-reader pipeline might be less efficient, since these two modules are not runnable in parallel.

An alternative solution tackling the limitations above is to directly predict the phrase-level answers consisting of a set of words, which are predicted from a set of documents in a large corpus. While this approach appears challenging, recent work shows that it is indeed possible to directly retrieve phrases within a text corpus based on their representational similarities to the input question~\cite{DBLP:conf/acl/SeoLKPFH19, DBLP:conf/acl/LeeSKC20, lee-2021-phrase}. However, its capability of retrieving phrases has been studied only with single-turn-based short questions, and their applications to ODConvQA, additionally requiring contextualizing the multi-turn conversations as well as effectively representing the lengthy conversational histories, have not been explored.

To this end, in this work, we first formulate the open-domain ConvQA task, previously done with the two-stage retriever-reader pipeline, as a direct phrase retrieval problem based on a single dense phrase retriever. However, in contrast to the single-turn open-domain question answering task that needs to understand only a single question, the target ODConvQA is more challenging since it has to comprehensively incorporate both the current question and the previous conversational histories in multi-turns.
For example, as shown in Figure~\ref{fig:0_concept}, in order to answer the question, `` What happened in 2003'', the model has to fully understand that the conversational context is related to the song, not the movie. 
While some work~\cite{DBLP:conf/sigir/Qu0CQCI20, DBLP:conf/ijcnlp/FangHHC22, DBLP:journals/tacl/AdlakhaDSVR22} proposes to feed an ODConvQA model the entire context consisting of the current question together with the conversational histories as an input, this naïve approach might be insufficient to solve the conversational dependency issue, which may lead to suboptimal performances in a phrase retrieval scheme.

In order to further address such a conversational dependency problem, we suggest to enforce the representation of the current conversational context to be similar to the representation of the previous context. 
Then, since two consecutive turns in a conversation are dependently represented in a similar embedding space, phrases that are relevant to both the current and previous conversational contexts are more likely to be retrieved, for the current question. To realize this objective, we maximize the representational similarities between the current conversational context and its previous contexts, while minimizing the representations between the current and its irrelevant contexts within the same batch via the contrastive learning loss, which is jointly trained with the dense phrase retriever. This is illustrated in Figure~\ref{fig:0_concept}, where we force the representation of the current conversational turn to be similar to its previous turn.
We refer to our proposed method as \textbf{P}hrase \textbf{R}etrieval for \textbf{O}pen\textbf{-}domain \textbf{Conv}ersational \textbf{Q}uestion \textbf{A}nswering (\textbf{PRO-ConvQA}).

We validate our proposed PRO-ConvQA method on two standard ODConvQA datasets, namely OR-QuAC~\cite{DBLP:conf/sigir/Qu0CQCI20} and TopiOCQA~\cite{DBLP:journals/tacl/AdlakhaDSVR22}, against diverse ODConvQA baselines that rely on the retriever-reader pipeline. The experimental results show that our PRO-ConvQA significantly outperforms relevant baselines. Furthermore, a detailed analysis demonstrates the effectiveness of the proposed contrastive learning strategy and the efficiency of our phrase retrieval strategy. Our contributions in this work are threefold:

\begin{itemize}[itemsep=1.0mm, parsep=1pt]
\vspace{-0.05in}
  \item We formulate a challenging open-domain conversational question answering (ODConvQA) problem into a dense phrase retrieval problem for the first time, by simplifying the conventional two-stage pipeline approach to ODConvQA tasks consisting of the retriever and the reader into one single phrase retriever. 
  \item We ensure that, when retrieving phrases, the representation for the current conversational context is similar to the representations for previous conversation histories, by modeling their conversational dependencies based on the contrastive learning strategy.
  \item We show that our PRO-ConvQA method achieves outstanding performances on two benchmark ODConvQA datasets against relevant baselines that use a pipeline approach.
\end{itemize}

\section{Related Work}

\paragraph{Conversational Question Answering}
ConvQA is similar to the reading comprehension task~\cite{squad, newsqa} in that it also aims at correctly answering the question from the given reference passage~\cite{DBLP:conf/emnlp/ChoiHIYYCLZ18,DBLP:journals/tacl/ReddyCM19}. However, ConvQA is a more difficult task than the reading comprehension task, since ConvQA has to answer questions interactively with users through multi-turns, which requires capturing all the contexts including previous conversational turns and the current question as well as its reference passage.  
To consider this unique characteristics, a line of research on ConvQA has focused on selecting only the query-relevant conversation history~\cite{DBLP:conf/iclr/HuangCY19, DBLP:conf/cikm/QuYQZCCI19, DBLP:conf/ijcai/0022WZ20, DBLP:conf/aaai/QiuHCJQ0HZ21}. 
However, recent work observed that a simple concatenation of the conversational histories outperforms the previous history selection approaches, thanks to the efficacy of the pre-trained language models~\cite{DBLP:conf/nips/VaswaniSPUJGKP17} in contextualizing long texts~\cite{DBLP:conf/acl/KimKPK20}. 
However, as the conversations often involve linguistic characteristics such as anaphora and ellipsis~\cite{DBLP:journals/kais/ZaibZSMZ22}, some work suggested to rewrite the ambiguous questions to explicitly model them~\cite{DBLP:conf/acl/KimKPK20, DBLP:conf/wsdm/VakulenkoLTA21, DBLP:conf/ecir/RaposoRMC22}. 
However, a naïve ConvQA setting assumes a fundamentally unrealistic setting, where the gold reference passages, containing answers corresponding to the questions, are already given.

\paragraph{Open-Domain ConvQA}
In order to address the unrealistic nature of the aforementioned ConvQA scenario, some recent work proposed to extend it to the open-retrieval scenario, which aims at retrieving relevant passages in response to the ongoing conversation and then uses them as reference passages, instead of using human-labeled passages. In this setting, effectively incorporating the conversational histories into the retrieval models is one of the main challenges, and several work~\cite{DBLP:conf/emnlp/LinYL21, DBLP:conf/sigir/YuLXF021, DBLP:conf/sigir/MaoDQ22, https://doi.org/10.48550/arxiv.2112.08558} proposed improving the first-stage retrievers, which are trained with particular machine learning techniques such as knowledge distillation, data augmentation, and reinforcement learning. However, their main focus is only on the first-stage retrieval aiming at returning only the query-related candidate passages, without giving exact answers to the questions. Also, some methods, such as ConvDR~\cite{DBLP:conf/sigir/YuLXF021} and ConvADR-QA~\cite{DBLP:conf/ijcnlp/FangHHC22}, use additional questions, which are rewritten from original questions by humans, to improve a retrieval performance by distilling the knowledge from the rewritten queries to the original queries. However, manually-rewritten queries are usually not available, and annotating them requires significant costs; therefore, they are trainable only under specific circumstances. On the other hand, to provide exact answers for the question within the current conversation turn, some other work adapted a retriever-reader pipeline, which can additionally read the query-relevant passages retrieved from a large corpus~\cite{DBLP:conf/sigir/Qu0CQCI20, DBLP:journals/tois/LiLN22, DBLP:journals/tacl/AdlakhaDSVR22, DBLP:conf/ijcnlp/FangHHC22}. However, such a pipeline approach has critical drawbacks due to its structural limitation composed of two sub-modules, thereby requiring additional effort to independently train both the retriever and the reader, both of which are also not runnable in parallel during inference, as well as bounding the reader's performance to the previous retrieval performance.

\paragraph{Dense Phrase Retrieval}
Instead of using a conventional pipeline approach, consisting of the retriever and the reader, we propose to directly predict answers for the ODConvQA task based on dense phrase retrieval. Following this line of previous researches, there exists some work that proposed to directly retrieve phrase-level answers from a large corpus; however, such work mainly focuses on non-conversational domains, such as question answering and relation extraction tasks~\cite{DBLP:conf/acl/SeoLKPFH19, DBLP:conf/acl/LeeSKC20, lee-2021-phrase}. Specifically, the pioneering work~\cite{DBLP:conf/acl/SeoLKPFH19} used both of the sparse and dense phrase representations for their retrieval. Afterwards,~\citet{DBLP:conf/acl/LeeSKC20} improved the phrase retrieval model that uses only dense representations without using any sparse representations, resulting in improved performance while reducing the memory footprint. Motivated by its effectiveness and efficiency, several work recently proposed to use the dense phrase retrieval system in diverse open-retrieval problems~\cite{lee-2021-phrase, DBLP:conf/acl/LiSM22, DBLP:journals/corr/abs-2112-08808}; however, their applicability to our target ODConvQA has been largely underexplored. Therefore, in this work, we adapt dense phrase retrieval to the ODConvQA task for the first time, and further propose to model conversational dependencies in phrase retrieval. 
\section{Method}
In this section, we first define the Conversational Question Answering (ConvQA) task, and its extension to the open-domain setting: Open-Domain ConvQA (ODConvQA) in Section~\ref{subsec:preliminaries}. Then, we introduce our dense phrase retrieval mechanism to effectively and efficiently solve the ODConvQA task, compared to the conventional retriever-reader pipeline approach, in Section~\ref{subsec:phraseretrieval}. Last, we explain our novel conversational dependency modeling strategy via contrastive learning, in Section~\ref{subsec:contrastive}.

\subsection{Preliminaries}
\label{subsec:preliminaries}
In this subsection, we first provide general descriptions of the ConvQA and the ODConvQA tasks.

\paragraph{Conversational Question Answering}
Let $q_i$ be the question and $a_i$ be the answer for the $i$-th turn of the conversation. Also, let $p^{*}_i$ a reference passage, which contains the answer $a_i$ for the question $q_i$. Then, given $q_i$, the goal of the ConvQA task is to correctly predict the answer $a_i$ based on the reference passage $p^{*}_i$ and the previous conversation histories: $\{q_{i-1}, a_{i-1}, ..., q_1, a_1\}$. Here, for the simplicity of the notation, we denote the $i$-th conversational context as the concatenation of the current input question and the previous conversation histories, formally represented as follows:
\begin{equation}
   \texttt{Conv}_i = \{q_i, q_{i-1}, a_{i-1}, ..., q_1, a_1\}.
\label{eq:conv_turn}
\end{equation}

Then, based on the notation of the conversational context $\texttt{Conv}_i$, we formulate the objective of the ConvQA task with a scoring function $f$, as follows:
\begin{equation}
    f(a_i|\texttt{Conv}_i) = M_{cqa} (p^{*}_i, \texttt{Conv}_i; \theta_{cqa}),
\label{eq:convqa_objective}
\end{equation}
where $M_{cqa}$ is a certain ConvQA model that predicts $a_i$ from $p^{*}_i$ based on $\texttt{Conv}_i$, which is parameterized by $\theta_{cqa}$. However, this setting of providing the reference passage $p^{*}_i$ containing the exact answer $a_i$ is largely unrealistic, since such the gold passage is usually not available when conversing with users in the real-world scenario. Therefore, in this work, we consider the more challenging open-domain ConvQA scenario, where we should extract the answers within the query-related documents from a large corpus, such as Wikipedia.  

\paragraph{Open-Domain ConvQA}
Unlike the ConvQA task that aims at extracting the answers from the gold passage $p^{*}_i$, the ODConvQA task is required to search a collection of passages for the relevant passages and then extract answers from them. Therefore, the scoring function $f$ of the ODConvQA task is formulated along with the certain passage $p_j$ from the large corpus $\mathcal{P}$, as follows:
\begin{equation}
\begin{split}
    f(a_i|\texttt{Conv}_i) = \; & M_{odcqa} (p_j, \texttt{Conv}_i; \theta_{odcqa}), \\
    & \text{with} \; p_j \in \mathcal{P},
\label{eq:orconvqa_objective}
\end{split}
\end{equation}
where $M_{odcqa}$ is an ODConvQA model parameterized by $\theta_{odcqa}$, and $\mathcal{P}$ is a collection of passages.

\paragraph{Retriever-Reader}
To realize the scoring function in Equation~\ref{eq:orconvqa_objective} for ODConvQA, the retriever-reader pipeline approach is dominantly used, which first retrieves the top-$K$ query-relevant passages and then reads a set of retrieved passages to answer the question based on them. Therefore, for this pipeline approach, the scoring function $f$ is decomposed into two sub-components (i.e., retriever and reader), formally defined as follows:
\begin{equation}
\begin{split}
   f(a_i|\texttt{Conv}_i) & = M_{retr}(\mathcal{P}_K | \texttt{Conv}_i;\theta_{retr}) \\ & \; \times M_{read}(a_i | \mathcal{P}_K ;\theta_{read}), 
\label{eq:retriever_reader_scoring}
\end{split}
\end{equation}
where the first-stage retriever $M_{retr}$ and the second-stage reader $M_{read}$ are parameterized with $\theta_{retr}$ and $\theta_{read}$, respectively. Also, $\mathcal{P}_K$ indicates a set of top-$K$ query-relevant passages, which are retrieved from the large corpus, $\mathcal{P}_K \subset \mathcal{P}$, based on the retriever $M_{retr}$.
However, such a retriever-reader pipeline is problematic for the following reasons. First, it is prone to error propagation from the retriever to the reader, since, if $M_{retr}$ retrieves irrelevant passages $\mathcal{P}_K$ that do not contain the answer such that $a_i \notin \mathcal{P}_K$, the reader $M_{read}$ fails to answer correctly. Second, it is inefficient, since $M_{read}$ requires the $M_{retr}$'s output as the input; therefore, $M_{retr}$ and $M_{read}$ are not runnable in parallel. Last, it demands effort to construct both $M_{retr}$ and $M_{read}$.

\subsection{Dense Phrase Retrieval for ODConvQA}
\label{subsec:phraseretrieval}
In order to address the aforementioned limitations of the retriever-reader pipeline for ODConvQA, in this work, we newly formulate the ODConvQA task as a dense phrase retrieval problem. In other words, we aim at directly retrieving the answer $a_i$, consisting of a sequence of words (i.e., phrase), based on its representational similarity to the conversational context $\texttt{Conv}_i$ via the dense phrase retriever~\cite{DBLP:conf/acl/LeeSKC20}. Formally, the scoring function for our ODConvQA based on the phrase retrieval scheme is defined as follows: 
\begin{equation}
   f(a_i|\texttt{Conv}_i) = E_{ConvQ}(\texttt{Conv}_i)^\top E_{A}(a_i), 
\label{eq:dense_phrase_scoring}
\end{equation}
where $E_{ConvQ}$ and $E_{A}$ are encoders that represent the conversational context $\texttt{Conv}_i$ and the phrase-level answer $a_i$, respectively. Also, $^\top$ symbol denotes inner product between its left and right terms. We note that this phrase retrieval mechanism defined in Equation~\ref{eq:dense_phrase_scoring} is similarly understood as predicting the answer in the reading comprehension task~\cite{squad, bidaf}. To be specific, in the reading comprehension task, we predict the start and end tokens of the answer $a_i$ located in the gold passage $p^*_i$. Similarly, in the phrase retrieval task, we directly predict the start and end tokens of the answer which is located within one part of the entire total passages $\mathcal{P}$; therefore, all words in all passages are sequentially pre-indexed and the goal is to find only the locations of the answer based on its similarity to the input context, e.g., $\texttt{Conv}_i$. Note that this phrase retrieval approach simplifies the conventional two-stage pipeline approach, commonly used for ODConvQA tasks, into the single direct answer retrieval, by removing the phrase reading done over the retrieved documents.

The training objective of the most information retrieval work~\cite{DBLP:conf/emnlp/KarpukhinOMLWEC20, qu-etal-2021-rocketqa} is to rank the pair of the query and its relevant documents highest among all the other irrelevant pairs. Similar to this, our training objective with a dense phrase retriever is formalized as follows: 
\begin{equation}
   \mathcal{L}_{neg} = -\log \cfrac{e^{f(a^+, \texttt{Conv}_i)}}{e^{f(a^+, \texttt{Conv}_i)} + \sum\limits_{k=1}^{N} e^{f({a}^-, \texttt{Conv}_i)}},
\label{eq:contrastive}
\end{equation}
where, for the context $\texttt{Conv}_i$, $a^+$ is the positive answer phrase and ${a}^-$ is the negative answer phrase. We describe how to construct the negative context-phrase pairs and additional details for training of the dense phrase retriever in the paragraph below.

\paragraph{Training Details}
In order to improve the performance of the dense phrase retriever, we adopt the existing strategies following~\citet{DBLP:conf/acl/LeeSKC20}.
First of all, we construct the negative samples, used in Equation~\ref{eq:contrastive}, based on in-batch and pre-batch sampling strategy. Specifically, for the $B$ number of phrases in the batch, $(B-1)$ in-batch phrases are used for negative samples by excluding one positive phrase with regard to the certain conversation context. Also, given the preceding $C$ number of batches, we can obtain the negative phrases for the current conversation context with a size of $(B \times C)$. In addition to negative sampling, we use the query-side fine-tuning scheme, which optimizes only the conversational question encoder, $E_{ConvQ}$, by maximizing the representational similarities between the correctly retrieved phrases and their corresponding conversational contexts after the phrase indexing. Last, to further improve predicting the start and end spans of the phrase retriever, we first train the reading comprehension model and then distill its knowledge, by minimizing the KL divergences of span predictions between the reading comprehension model and the phrase retriever. For more details, please refer to~\citet{DBLP:conf/acl/LeeSKC20}.

\subsection{Conversational Dependency Modeling}
\label{subsec:contrastive}
While Equation~\ref{eq:contrastive} effectively discriminates positive answer phrases from negative answer phrases, relying on it is sub-optimal when solving the ODConvQA task, where each conversational turn shares a similar context with its previous turn. 
In other words, since information-seeking conversational questions are asked in a sequence, two consecutive contexts, $\texttt{Conv}_{i-1}$ and $\texttt{Conv}_{i}$, should have similar representations compared to the other turns from different conversations. Therefore, we further model such a conversational dependency by maximizing the similarity between the sequential turns while minimizing the similarity between the other irrelevant turns via contrastive learning as follows:
 \begin{equation}
 \fontsize{9.25pt}{9.25pt}\selectfont
   \mathcal{L}_{turn} = -\log \cfrac{e^{f(\texttt{Conv}_i, \texttt{Conv}_{i-1})}}{e^{f(\texttt{Conv}_i, \texttt{Conv}_{i-1})} + \sum\limits_{k=1}^{B-1} e^{f({\texttt{Conv}_i}^-, \texttt{Conv}_{i-1})}},
\label{eq:history_contrastive}
\fontsize{9.25pt}{9.25pt}\selectfont
\end{equation}
where ${\texttt{Conv}_i}^-$ comes from a collection of the irrelevant conversation turns within the batch. By optimizing the objective in Equation~\ref{eq:history_contrastive}, the encoder $E_{ConvQ}$ represents the current conversational turn $\texttt{Conv}_i$ probably similar to its previous turn $\texttt{Conv}_{i-1}$; therefore, the retrieved phrase captures both the current and previous conversational contexts.

\paragraph{Overall Training objective}
We optimize the phrase retrieval loss from Equation~\ref{eq:contrastive} and conversational dependency loss from Equation~\ref{eq:history_contrastive} as follows:
\begin{equation}
   \mathcal{L} = \lambda_{1} \mathcal{L}_{neg} + \lambda_{2} \mathcal{L}_{turn},
\label{eq:final_loss}
\end{equation}
where $\lambda_{1}$ and $\lambda_{2}$ are the weights for each loss term.

\section{Experimental Setups}
In this section, we explain datasets, metrics, models, and implementation details.

\subsection{Datasets and Metrics}
\paragraph{OR-QuAC}
OR-QuAC~\cite{DBLP:conf/sigir/Qu0CQCI20} is the benchmark ODConvQA dataset, which extends a popular ConvQA dataset, namely QuAC~\cite{DBLP:conf/emnlp/ChoiHIYYCLZ18}, to the open-retrieval setting. This dataset consists of 35,526 conversational turns for training, 3,430 for validation, and 5,571 for testing. 

\paragraph{TopiOCQA}
TopiOCQA~\cite{DBLP:journals/tacl/AdlakhaDSVR22} is another ODConvQA dataset that considers the topic-switching problem across different conversational turns. This dataset contains 45,450 conversational turns and 2,514 turns for training and validation, respectively. Note that we use a validation set since the test set is not publicly open. 

\paragraph{Evaluation Metrics}
We evaluate all models with F1-score and extact match (EM) following the standard protocol on the ODConvQA tasks~\cite{DBLP:conf/sigir/Qu0CQCI20, DBLP:journals/tacl/AdlakhaDSVR22}. Also, for retrieval performances, we use the standard ranking metrics: Top-K accuracy, mean reciprocal rank (MRR), and Precision, following~\citet{lee-2021-phrase}.

\subsection{Baselines and Our Model}
We introduce the baselines with a retriever-reader pipeline, which is dominantly adopted for ODConvQA. We do not compare against the incomparable baselines that use the additional data, such as rewritten queries~\cite{DBLP:conf/sigir/YuLXF021, DBLP:conf/ijcnlp/FangHHC22}.

\paragraph{BM25 Retriever + DPR Reader}
This is one of the most widely used retriever-reader pipeline approaches that first retrieves query-relevant passages with a sparse retriever, BM25~\cite{DBLP:conf/trec/RobertsonWJHG94}, and then reads top-$k$ retrieved passages with a DPR reader~\cite{DBLP:conf/emnlp/KarpukhinOMLWEC20}.
 
\paragraph{DPR Retriever + DPR Reader}
This pipeline uses a dense retriever for the first retrieval stage, DPR retriever~\cite{DBLP:conf/emnlp/KarpukhinOMLWEC20}, which calculates the similarity between a query and passages on a latent space, instead of using a sparse retriever.

\paragraph{ORConvQA}
This model consists of a dense retriever and a reader with an additional re-ranker, which is trained with two phases~\cite{DBLP:conf/sigir/Qu0CQCI20}: 1) retriever pre-training and 2) concurrent learning. Specifically, it first trains the retriever and generates dense passage representations. Then, the model further trains the retriever, reader, and re-ranker using the pre-trained retriever and generated passage representations.

\paragraph{PRO-ConvQA(Ours)}
This is our model that directly retrieves answers without passage reading, trained jointly with contrastive learning to further address a conversational dependency issue.

\aboverulesep=0ex
\belowrulesep=0ex

\begin{table}
\centering
\begin{center}

\resizebox{0.49\textwidth}{!}{
\begin{tabular}[h]{l|cccc}
\toprule
\multicolumn{1}{c|}{} & \multicolumn{2}{c}{\textbf{OR-QuAC}} & \multicolumn{2}{c}{\textbf{TopiOCQA}} \\
\cmidrule(l{2pt}r{2pt}){2-3} \cmidrule(l{2pt}r{2pt}){4-5} 
\multicolumn{1}{c|}{} & \textbf{F1} & \textbf{EM} & \textbf{F1} & \textbf{EM} \\
\midrule
\multirow{1}{*}{BM25 Ret. + DPR Read.}
 & 30.82 & 11.17 & 13.92 & 4.09 \\
\multirow{1}{*}{DPR Ret. + DPR Read.}
 & 25.94  & 8.15 & 23.13 & 9.06 \\
\multirow{1}{*}{ORConvQA}
 & 28.86 & 14.39 & 10.67 & 2.36 \\
 \hline
 \multirow{1}{*}{PRO-ConvQA (Ours)}
 & \textbf{36.84} &  \textbf{15.73} & \textbf{36.67} & \textbf{20.38} \\
\bottomrule
\end{tabular}
}
\end{center}
\vspace{-0.13in}
\caption{\small F1 and EM scores on OR-QuAC and TopioCQA. Note that the best scores are highlighted in \textbf{bold}. 
}
\label{tab:main}
\vspace{-0.1in}
\end{table}

\subsection{Implementation Details}
We implement ODConvQA models using PyTorch~\cite{DBLP:conf/nips/PaszkeGMLBCKLGA19} and Transformers library~\cite{DBLP:conf/emnlp/WolfDSCDMCRLFDS20}.
For all the models, we use the 2018-12-20 Wikipedia snapshot having a collection of 16,766,529 passages. We exclude the questions with unanswerable answers, since we cannot find their answers with the corpus, which is not suitable for the goal of the open-retrieval problem. Furthermore, as our model answers questions extractively, we convert TopiOCQA with the gold answers in a free-form text to our extractive setting by considering the provided rationale as the gold answers, following the existing setting from~\citet{jeong-etal-2023-realistic}.
For training PRO-ConvQA, we set the batch size ($B$) as $24$ and the pre-batch size ($C$) as $2$. Also, We train PRO-ConvQA with 3 epochs with a learning rate of $3e-5$ and further fine-tune a query encoder with 3 epochs. We set $\lambda_{1}$ and $\lambda_{2}$ as 4 and 1 for OR-QuAC and 2 and 1 for TopiOCQA, respectively.
For computing resources, we use two GeForce RTX 3090 GPUs with 24GB memory. 
For retriever-reader baselines, we retrieve top-$5$ passages to train and evaluate the reader, following~\citet{DBLP:conf/sigir/Qu0CQCI20}. Also, due to the significant costs of evaluating retrieval models, we perform experiments with a single run.

\section{Results and Discussion}
\begin{figure}[t!]
\begin{center}
\includegraphics[width=0.49\textwidth]{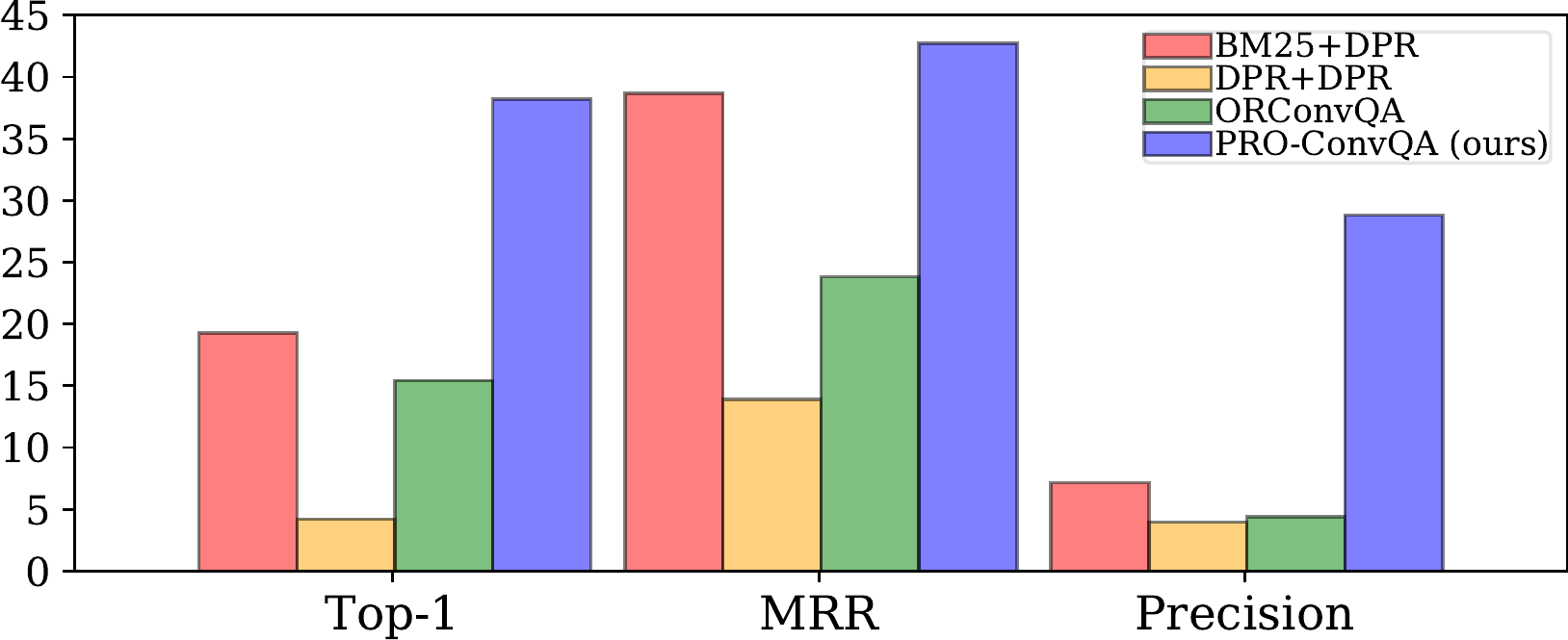}
\end{center}
\vspace{-0.13in}
\caption{
\small Retrieval results on OR-QuAC, measured with Top-$1$ accuracy, MRR, and Precision. Note that we limit the number of total retrieved documents for MRR and Precision to $10$.
}
\vspace{-0.1in}
\label{fig:mrr}
\end{figure}

In this section, we show the overall results and provide detailed analyses.

\paragraph{Main Results}

As Table~\ref{tab:main} shows, our proposed PRO-ConvQA model significantly outperforms all baselines with a retriever-reader pipeline on two benchmark datasets. This implies that the two-stage models might be susceptible to error propagation between the retrieval and reader stages, therefore ineffectively bounding the overall performances when a model fails to correctly retrieve reference passages during the first stage. However, our PRO-ConvQA is free from such a bottleneck problem, since it directly retrieves answer phrases, without requiring an additional reader. 

Interestingly, a recent ORConvQA model shows largely inferior performances on the TopiOCQA dataset. Note that for TopiOCQA, target passages of two consecutive conversation turns sometimes have different topics, compared to the OR-QuAC dataset where all passages within the whole conversation share a single topic. Therefore, TopiOCQA follows a more realistic setting where a topic constantly changes during the conversation. However, note that ORConvQA is not trained in a truly end-to-end fashion, since it first retrieves passage embeddings from a pre-trained retriever, and then uses the already encoded passage embeddings when concurrently training a retriever, reader, and re-ranker. Therefore, ORConvQA is vulnerable to such a topic-shifting situation, as the passage encoder and embedding are not updated during a concurrent training step. Meanwhile, our PRO-ConvQA is trained in an end-to-end fashion, thereby effectively learning to retrieve phrases. 

Similarly, using BM25 as a first-stage retriever also shows a large performance gap between the two datasets. Note that BM25 lexically measures relevance between a conversational turn and a passage by counting their overlapping terms. Therefore, compared to the other dense-retrieval-based two-stage models, this unique characteristic of BM25 brings additional advantages on the OR-QuAC dataset, where each conversational turn revolves around the same topic. More specifically, the conversational history, which is accumulated during each turn, becomes very relevant to the target retrieval passage as the conversation progresses. However, such a lexical comparison scheme fails to effectively retrieve the passages when a topic slightly changes for each conversation turn on TopiOCQA, since it cannot capture a semantic inter-relationship between conversational turns and a passage. On the other hand, our PRO-ConvQA shows robust performances on both datasets by retrieving the phrases over the semantic representation space. We further analyze the strengths of the PRO-ConvQA in the following paragraphs.

\aboverulesep=0ex
\belowrulesep=0ex

\begin{table}[t]
\centering
\begin{center}
\resizebox{0.49\textwidth}{!}{
\begin{tabular}[t]{l|cc}
\toprule
\multicolumn{1}{c|}{} &\textbf{Relative Time} &\textbf{\#Q / sec.} \\
\midrule
\multirow{1}{*}{BM25 Ret. + DPR Read.}
 & 16.94 & 1.74 \\
\multirow{1}{*}{DPR Ret. + DPR Read.}
 & 15.48 & 1.91 \\
 \multirow{1}{*}{ORConvQA}
 & 10.95 & 2.70 \\
 \hline
 \multirow{1}{*}{PRO-ConvQA (Ours)}
 & \textbf{1.00} & \textbf{29.6}  \\
\bottomrule
\end{tabular}
}
\end{center}
\vspace{0.07in}
\caption{\small Wall-clock time for inference on TopiOCQA. Note that we measure the total inference time required to output an answer, thereby considering both retrieving and reading time.}
\label{tab:efficiency}
\vspace{-0.1in}
\end{table}

\paragraph{Effectiveness on Retrieval Performance}
In order to validate whether a failure of the retriever works as a bottleneck in a two-stage pipeline, we measure retrieval performances in Figure~\ref{fig:mrr}. Compared to the PRO-ConvQA, the models based on the retriever-reader pipeline fail to correctly retrieve relevant reference passages, thus negatively leading to the degenerated overall performance. This result corroborates our hypothesis that there exists a bottleneck problem in the first retrieval stage. Furthermore, this result demonstrates that our PRO-ConvQA also effectively retrieves the related passages at a phrase level, even though it is not directly designed to solve the conversational search task that aims at only retrieving the passages related to each conversational turn.

\paragraph{Efficiency on Inference Time}
In the real world, inference speed for returning answers to the given questions is crucially important. Thus, we report the runtime efficiency of our PRO-ConvQA against the other baselines in Table~\ref{tab:efficiency}. Note that PRO-ConvQA is highly efficient for searching answer phrases over the baselines with a retriever-reader pipeline. This is because retrieval and reader stages cannot be run in parallel, since the latter reader stage requires the retrieved passages as the input. On the other hand, our proposed PRO-ConvQA is simply composed of a single phrase retrieval stage with two decomposable encoders, as formulated in Equation~\ref{eq:dense_phrase_scoring}. This decomposable feature enables maximum inner product search (MIPS), thus contributing to fast inference speed.

\aboverulesep=0ex
\belowrulesep=0ex

\begin{table}[t]
\centering
\begin{center}
\resizebox{0.49\textwidth}{!}{
\begin{tabular}[t]{l|cccc}
\toprule
\multicolumn{1}{c|}{} &\textbf{CL} &\textbf{QF} &\textbf{F1} &\textbf{EM}\\
\midrule
\multirow{1}{*}{PRO-ConvQA (Ours)}
 & \cmark & \cmark & \textbf{36.84} & \textbf{15.73} \\
\multirow{1}{*}{PRO-ConvQA w/o QF}
 & \cmark & \xmark & 33.00 & 13.07 \\
 \multirow{1}{*}{PRO-ConvQA w/o CL}
 & \xmark & \cmark & 33.53 & 13.20 \\
 \multirow{1}{*}{PRO-ConvQA w/o CL, QF}
 & \xmark & \xmark & 30.33 & 11.14 \\
\bottomrule
\end{tabular}
}
\end{center}
\vspace{-0.025in}
\caption{\small Ablation studies of our PRO-ConvQA on the OR-QuAC dataset. Note that CL and QF refer to contrastive learning and query-side fine-tuning strategies, respectively. }
\vspace{-0.1in}
\label{tab:ablation}
\end{table}

\paragraph{Ablation Studies}
To understand how each component in the PRO-ConvQA contributes to performance gains, we provide ablation studies in Table~\ref{tab:ablation}. As shown in Table~\ref{tab:ablation}, our contrastive learning for conversational dependency modeling and also query-side fine-tuning strategies positively contribute to the overall performance. Furthermore, the significant performance drops when removing each component indicate that there exists a complementary relation between the two components.

\begin{figure}[t!]
\begin{center}
\includegraphics[width=0.49\textwidth]{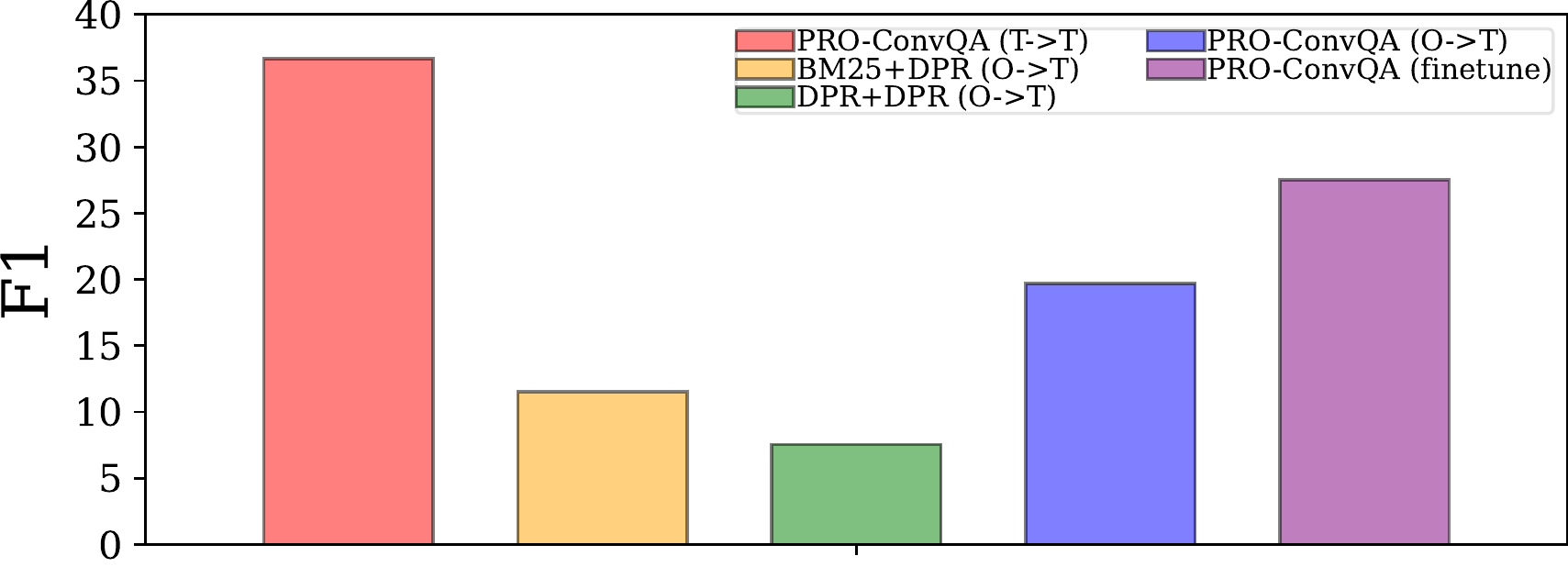}
\end{center}
\vspace{-0.15in}
\caption{
\small F1-scores in a zero-shot setting where a model is trained on OR-QuAC (O) and evaluated on TopiOCQA (T). Finetune denotes the query-side fine-tuning on TopiOCQA. 
}
\vspace{-0.1in}
\label{fig:zeroshot}
\end{figure}

\paragraph{Zero-shot Performance}
In order to apply ODConvQA models in a real-world scenario, one may consider a zero-shot performance since high-quality training data is not always available. Therefore, we show zero-shot performances, assuming that the target training data is only available for OR-QuAC, but not for TopiOCQA. As Figure~\ref{fig:zeroshot} shows, the proposed PRO-ConvQA outperforms the baseline models by a large margin. This implies that such a zero-shot setting is challenging to the previous ODConvQA models, since they are trained and tested in a different topic-shifting setting; they are trained to assume that each turn shares the same topic within a conversation, but tested in a situation where the topic changes as the conversation proceeds. However, PRO-ConvQA is more robust than other baselines in a zero-shot setting, since its training objective aims at retrieving answers at a phrase-level, rather than a passage-level, which enables capturing topic shifts with more flexibility.

\paragraph{Efficient Transfer Learning}
Besides a zero-shot performance, transferability between different datasets is another important feature to consider in a real-world scenario. In particular, it would be efficient to reuse a dump of phrase embeddings and indexes even if the target data changes, with respect to the training effort and disk footprint for storing a large size of embeddings and indexes.
As we have validated the effectiveness of fine-tuning a query encoder in Table~\ref{tab:ablation}, it would be more efficient if we could only update the query encoder to adapt to the newly given data, without re-training everything from scratch. To see this, we conduct an experiment in a transfer learning scenario, where a phrase retrieval model is trained on OR-QuAC, but the query-side encoder is further fine-tuned for TopiOCQA and tested on it. As Figure~\ref{fig:zeroshot} shows, fine-tuning a query-side encoder further improves the performance when compared to the zero-shot model. This indicates that PRO-ConvQA can be efficiently adapted to diverse realistic settings, only compensating a little amount of costs for adaption. 

\paragraph{Generative Reader}

\begin{figure}[t!]
\begin{center}
\includegraphics[width=0.49\textwidth]{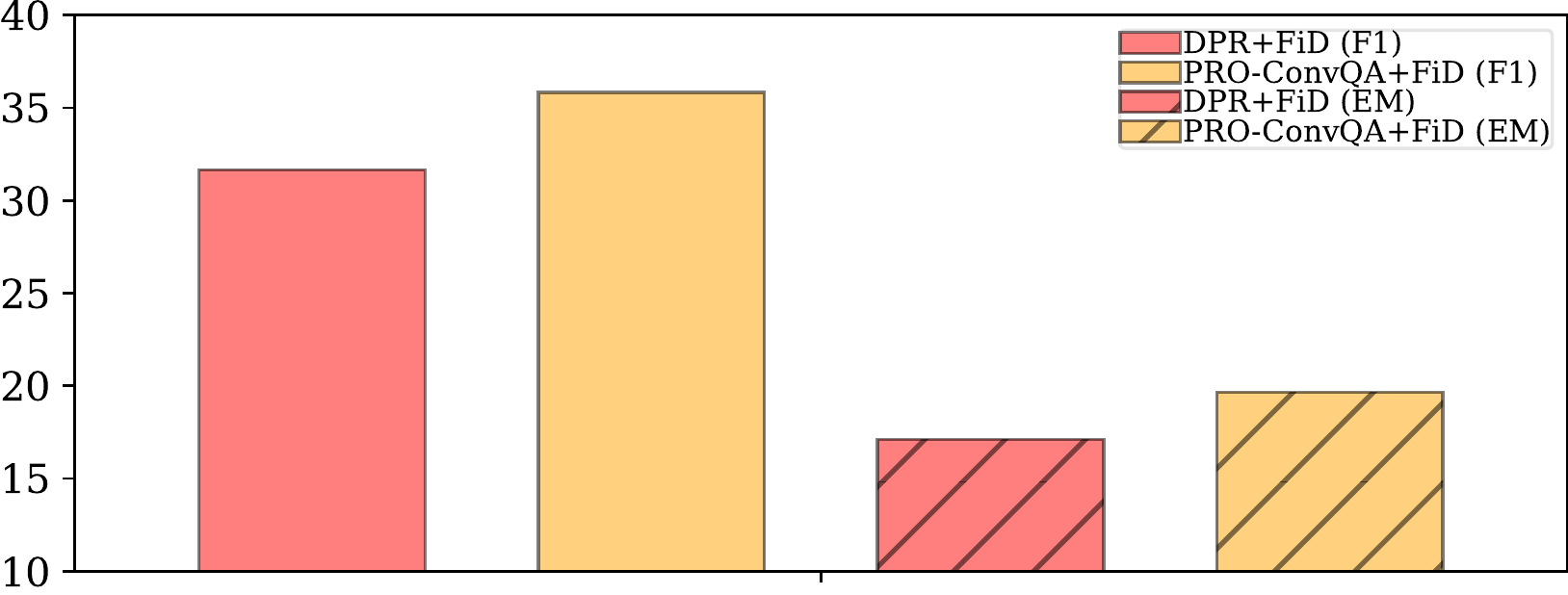}
\end{center}
\vspace{-0.13in}
\caption{
\small F1 and EM scores on TopiOCQA with a generative reader, namely FiD~\cite{DBLP:conf/eacl/IzacardG21}. 
}
\vspace{-0.1in}
\label{fig:fid}
\end{figure}

While our PRO-ConvQA shows outstanding performances under the extractive reader setting, it is also possible to further combine PRO-ConvQA with a recent generative reader model, Fusion-in-Decoder (FiD)~\cite{DBLP:conf/eacl/IzacardG21}. We conduct experiments with the publicly available FiD model\footnote{https://github.com/McGill-NLP/topiocqa}, which is already trained on TopiOCQA, without any further training. As Figure~\ref{fig:fid} shows, our PRO-ConvQA consistently shows superior F1 and EM scores under the generative reader setting, compared to the DPR baseline. This is because PRO-ConvQA is superior in passage-level retrieval as shown in Figure~\ref{fig:mrr}, which further leads to accurately answering questions with correctly retrieved passages. Also, we believe that the performance would be further improved by additionally training a FiD model on the retrieved passages from PRO-ConvQA, instead of using an already trained one.
\section{Conclusion}
In this work, we pointed out the limitations of the retriever-reader pipeline approach to ODConvQA, which is prone to error propagation from the retriever, unable to run both sub-modules in parallel, and demanding effort to manage these two sub-modules, due to its decomposed structure. To address such issues, we formulated the ODConvQA task as a dense phrase retrieval problem, which makes it possible to directly retrieve the answer based on its representational similarity to the current conversational context. Furthermore, to model the conversational dependency between the current and its previous turns, we force their representations to be similar with contrastive learning, which leads to retrieving more related phrases to the conversational history as well as the current question. We validated our proposed PRO-ConvQA on ODConvQA benchmark datasets, showing its efficacy in effectiveness and efficiency.

\section*{Limitations}
As shown in Table~\ref{tab:ablation}, the contrastive learning strategy to model the conversational dependencies between the current and previous conversational turns is a key element in our phrase retrieval-based ODConvQA task. However, when the current conversational topic is significantly shifted from the previous topic as the user may suddenly come up with new ideas, our contrastive learning strategy might be less effective. This is because modeling the conversational dependency is, in this case, no longer necessary. While we believe such situations are less frequent, one may further tackle this scenario of significant topic switching, for example, with history filtering, which we leave as future work.

\section*{Ethics Statement}
We show clear advantages of our PRO-ConvQA framework for ODConvQA tasks compared to the retriever-reader approach in both effectiveness and efficiency perspectives. However, when given the conversational context from malicious users who ask for offensive and harmful content, our PRO-ConvQA framework might become vulnerable to retrieving toxic phrases. Therefore, before deploying our PRO-ConvQA to real-world scenarios, we have to ensure the safety of the retrieved phrases.

\section*{Acknowledgements}
This work was supported by Institute for Information and communications Technology Promotion (IITP) grant funded by the Korea government (No. 2018-0-00582, Prediction and augmentation of the credibility distribution via linguistic analysis and automated evidence document collection).

\bibliography{anthology,custom}

\begin{thebibliography}{36}
\expandafter\ifx\csname natexlab\endcsname\relax\def\natexlab#1{#1}\fi

\bibitem[{Adlakha et~al.(2022)Adlakha, Dhuliawala, Suleman, de~Vries, and
  Reddy}]{DBLP:journals/tacl/AdlakhaDSVR22}
Vaibhav Adlakha, Shehzaad Dhuliawala, Kaheer Suleman, Harm de~Vries, and Siva
  Reddy. 2022.
\newblock \href {https://doi.org/10.1162/tacl\_a\_00471} {Topiocqa: Open-domain
  conversational question answering with topic switching}.
\newblock \emph{Trans. Assoc. Comput. Linguistics}, 10:468--483.

\bibitem[{Anantha et~al.(2021)Anantha, Vakulenko, Tu, Longpre, Pulman, and
  Chappidi}]{DBLP:conf/naacl/AnanthaVTLPC21}
Raviteja Anantha, Svitlana Vakulenko, Zhucheng Tu, Shayne Longpre, Stephen
  Pulman, and Srinivas Chappidi. 2021.
\newblock \href {https://doi.org/10.18653/v1/2021.naacl-main.44} {Open-domain
  question answering goes conversational via question rewriting}.
\newblock In \emph{Proceedings of the 2021 Conference of the North American
  Chapter of the Association for Computational Linguistics: Human Language
  Technologies, {NAACL-HLT} 2021}, pages 520--534. Association for
  Computational Linguistics.

\bibitem[{Chen et~al.(2020)Chen, Wu, and Zaki}]{DBLP:conf/ijcai/0022WZ20}
Yu~Chen, Lingfei Wu, and Mohammed~J. Zaki. 2020.
\newblock \href {https://doi.org/10.24963/ijcai.2020/171} {Graphflow:
  Exploiting conversation flow with graph neural networks for conversational
  machine comprehension}.
\newblock In \emph{Proceedings of the Twenty-Ninth International Joint
  Conference on Artificial Intelligence, {IJCAI} 2020}, pages 1230--1236.
  ijcai.org.

\bibitem[{Choi et~al.(2018)Choi, He, Iyyer, Yatskar, Yih, Choi, Liang, and
  Zettlemoyer}]{DBLP:conf/emnlp/ChoiHIYYCLZ18}
Eunsol Choi, He~He, Mohit Iyyer, Mark Yatskar, Wen{-}tau Yih, Yejin Choi, Percy
  Liang, and Luke Zettlemoyer. 2018.
\newblock \href {https://doi.org/10.18653/v1/d18-1241} {Quac: Question
  answering in context}.
\newblock In \emph{Proceedings of the 2018 Conference on Empirical Methods in
  Natural Language Processing}, pages 2174--2184. Association for Computational
  Linguistics.

\bibitem[{Fang et~al.(2022)Fang, Hung, Huang, and
  Chen}]{DBLP:conf/ijcnlp/FangHHC22}
Hung{-}Chieh Fang, Kuo{-}Han Hung, Chen{-}Wei Huang, and Yun{-}Nung Chen. 2022.
\newblock \href {https://aclanthology.org/2022.findings-aacl.30} {Open-domain
  conversational question answering with historical answers}.
\newblock In \emph{Findings of the Association for Computational Linguistics:
  {AACL-IJCNLP} 2022}, pages 319--326. Association for Computational
  Linguistics.

\bibitem[{Huang et~al.(2019)Huang, Choi, and Yih}]{DBLP:conf/iclr/HuangCY19}
Hsin{-}Yuan Huang, Eunsol Choi, and Wen{-}tau Yih. 2019.
\newblock \href {https://openreview.net/forum?id=ByftGnR9KX} {Flowqa: Grasping
  flow in history for conversational machine comprehension}.
\newblock In \emph{7th International Conference on Learning Representations,
  {ICLR} 2019, New Orleans, LA, USA, May 6-9, 2019}.

\bibitem[{Izacard and Grave(2021)}]{DBLP:conf/eacl/IzacardG21}
Gautier Izacard and Edouard Grave. 2021.
\newblock \href {https://doi.org/10.18653/v1/2021.eacl-main.74} {Leveraging
  passage retrieval with generative models for open domain question answering}.
\newblock In \emph{Proceedings of the 16th Conference of the European Chapter
  of the Association for Computational Linguistics: Main Volume, {EACL} 2021},
  pages 874--880. Association for Computational Linguistics.

\bibitem[{Jeong et~al.(2023)Jeong, Baek, Hwang, and
  Park}]{jeong-etal-2023-realistic}
Soyeong Jeong, Jinheon Baek, Sung~Ju Hwang, and Jong Park. 2023.
\newblock \href {https://aclanthology.org/2023.eacl-main.35} {Realistic
  conversational question answering with answer selection based on calibrated
  confidence and uncertainty measurement}.
\newblock In \emph{Proceedings of the 17th Conference of the European Chapter
  of the Association for Computational Linguistics}, pages 477--490, Dubrovnik,
  Croatia. Association for Computational Linguistics.

\bibitem[{Karpukhin et~al.(2020)Karpukhin, Oguz, Min, Lewis, Wu, Edunov, Chen,
  and Yih}]{DBLP:conf/emnlp/KarpukhinOMLWEC20}
Vladimir Karpukhin, Barlas Oguz, Sewon Min, Patrick S.~H. Lewis, Ledell Wu,
  Sergey Edunov, Danqi Chen, and Wen{-}tau Yih. 2020.
\newblock \href {https://doi.org/10.18653/v1/2020.emnlp-main.550} {Dense
  passage retrieval for open-domain question answering}.
\newblock In \emph{Proceedings of the 2020 Conference on Empirical Methods in
  Natural Language Processing, {EMNLP} 2020}, pages 6769--6781. Association for
  Computational Linguistics.

\bibitem[{Kim et~al.(2021)Kim, Kim, Park, and Kang}]{DBLP:conf/acl/KimKPK20}
Gangwoo Kim, Hyunjae Kim, Jungsoo Park, and Jaewoo Kang. 2021.
\newblock \href {https://doi.org/10.18653/v1/2021.acl-long.478} {Learn to
  resolve conversational dependency: {A} consistency training framework for
  conversational question answering}.
\newblock In \emph{Proceedings of the 59th Annual Meeting of the Association
  for Computational Linguistics and the 11th International Joint Conference on
  Natural Language Processing, {ACL/IJCNLP} 2021}, pages 6130--6141.
  Association for Computational Linguistics.

\bibitem[{Kim et~al.(2022)Kim, Yoo, Yoon, Lee, and
  Kang}]{DBLP:journals/corr/abs-2112-08808}
Hyunjae Kim, Jaehyo Yoo, Seunghyun Yoon, Jinhyuk Lee, and Jaewoo Kang. 2022.
\newblock \href {https://arxiv.org/abs/2112.08808} {Simple questions generate
  named entity recognition datasets}.
\newblock In \emph{Proceedings of the 2022 Conference on Empirical Methods in
  Natural Language Processing, {EMNLP} 2022}. Association for Computational
  Linguistics.

\bibitem[{Lee et~al.(2021{\natexlab{a}})Lee, Sung, Kang, and
  Chen}]{DBLP:conf/acl/LeeSKC20}
Jinhyuk Lee, Mujeen Sung, Jaewoo Kang, and Danqi Chen. 2021{\natexlab{a}}.
\newblock \href {https://doi.org/10.18653/v1/2021.acl-long.518} {Learning dense
  representations of phrases at scale}.
\newblock In \emph{Proceedings of the 59th Annual Meeting of the Association
  for Computational Linguistics and the 11th International Joint Conference on
  Natural Language Processing, {ACL/IJCNLP} 2021}, pages 6634--6647.
  Association for Computational Linguistics.

\bibitem[{Lee et~al.(2021{\natexlab{b}})Lee, Wettig, and
  Chen}]{lee-2021-phrase}
Jinhyuk Lee, Alexander Wettig, and Danqi Chen. 2021{\natexlab{b}}.
\newblock \href {https://doi.org/10.18653/v1/2021.emnlp-main.297} {Phrase
  retrieval learns passage retrieval, too}.
\newblock In \emph{Proceedings of the 2021 Conference on Empirical Methods in
  Natural Language Processing}, pages 3661--3672, Online and Punta Cana,
  Dominican Republic. Association for Computational Linguistics.

\bibitem[{Li et~al.(2022{\natexlab{a}})Li, Gao, Goenka, and
  Chen}]{DBLP:conf/acl/LiGGC22}
Huihan Li, Tianyu Gao, Manan Goenka, and Danqi Chen. 2022{\natexlab{a}}.
\newblock \href {https://doi.org/10.18653/v1/2022.acl-long.555} {Ditch the gold
  standard: Re-evaluating conversational question answering}.
\newblock In \emph{Proceedings of the 60th Annual Meeting of the Association
  for Computational Linguistics, {ACL} 2022}, pages 8074--8085. Association for
  Computational Linguistics.

\bibitem[{Li et~al.(2022{\natexlab{b}})Li, Shang, and
  McAuley}]{DBLP:conf/acl/LiSM22}
Jiacheng Li, Jingbo Shang, and Julian~J. McAuley. 2022{\natexlab{b}}.
\newblock \href {https://doi.org/10.18653/v1/2022.acl-long.426} {Uctopic:
  Unsupervised contrastive learning for phrase representations and topic
  mining}.
\newblock In \emph{Proceedings of the 60th Annual Meeting of the Association
  for Computational Linguistics, {ACL} 2022}, pages 6159--6169. Association for
  Computational Linguistics.

\bibitem[{Li et~al.(2022{\natexlab{c}})Li, Li, and
  Nie}]{DBLP:journals/tois/LiLN22}
Yongqi Li, Wenjie Li, and Liqiang Nie. 2022{\natexlab{c}}.
\newblock \href {https://doi.org/10.1145/3498557} {Dynamic graph reasoning for
  conversational open-domain question answering}.
\newblock \emph{{ACM} Trans. Inf. Syst.}, 40(4):82:1--82:24.

\bibitem[{Lin et~al.(2021)Lin, Yang, and Lin}]{DBLP:conf/emnlp/LinYL21}
Sheng{-}Chieh Lin, Jheng{-}Hong Yang, and Jimmy Lin. 2021.
\newblock \href {https://doi.org/10.18653/v1/2021.emnlp-main.77}
  {Contextualized query embeddings for conversational search}.
\newblock In \emph{Proceedings of the 2021 Conference on Empirical Methods in
  Natural Language Processing, {EMNLP} 2021}, pages 1004--1015. Association for
  Computational Linguistics.

\bibitem[{Mao et~al.(2022)Mao, Dou, and Qian}]{DBLP:conf/sigir/MaoDQ22}
Kelong Mao, Zhicheng Dou, and Hongjin Qian. 2022.
\newblock \href {https://doi.org/10.1145/3477495.3531961} {Curriculum
  contrastive context denoising for few-shot conversational dense retrieval}.
\newblock In \emph{{SIGIR} '22: The 45th International {ACM} {SIGIR} Conference
  on Research and Development in Information Retrieval}, pages 176--186. {ACM}.

\bibitem[{Paszke et~al.(2019)Paszke, Gross, Massa, Lerer, Bradbury, Chanan,
  Killeen, Lin, Gimelshein, Antiga, Desmaison, K{\"{o}}pf, Yang, DeVito,
  Raison, Tejani, Chilamkurthy, Steiner, Fang, Bai, and
  Chintala}]{DBLP:conf/nips/PaszkeGMLBCKLGA19}
Adam Paszke, Sam Gross, Francisco Massa, Adam Lerer, James Bradbury, Gregory
  Chanan, Trevor Killeen, Zeming Lin, Natalia Gimelshein, Luca Antiga, Alban
  Desmaison, Andreas K{\"{o}}pf, Edward~Z. Yang, Zachary DeVito, Martin Raison,
  Alykhan Tejani, Sasank Chilamkurthy, Benoit Steiner, Lu~Fang, Junjie Bai, and
  Soumith Chintala. 2019.
\newblock \href
  {https://proceedings.neurips.cc/paper/2019/hash/bdbca288fee7f92f2bfa9f7012727740-Abstract.html}
  {Pytorch: An imperative style, high-performance deep learning library}.
\newblock In \emph{Advances in Neural Information Processing Systems 32: Annual
  Conference on Neural Information Processing Systems 2019}, pages 8024--8035.

\bibitem[{Qiu et~al.(2021)Qiu, Huang, Chen, Ji, Qu, Wei, Huang, and
  Zhang}]{DBLP:conf/aaai/QiuHCJQ0HZ21}
Minghui Qiu, Xinjing Huang, Cen Chen, Feng Ji, Chen Qu, Wei Wei, Jun Huang, and
  Yin Zhang. 2021.
\newblock \href {https://ojs.aaai.org/index.php/AAAI/article/view/17617}
  {Reinforced history backtracking for conversational question answering}.
\newblock In \emph{Thirty-Fifth {AAAI} Conference on Artificial Intelligence,
  {AAAI} 2021, Thirty-Third Conference on Innovative Applications of Artificial
  Intelligence, {IAAI} 2021, The Eleventh Symposium on Educational Advances in
  Artificial Intelligence, {EAAI} 2021}, pages 13718--13726. {AAAI} Press.

\bibitem[{Qu et~al.(2020)Qu, Yang, Chen, Qiu, Croft, and
  Iyyer}]{DBLP:conf/sigir/Qu0CQCI20}
Chen Qu, Liu Yang, Cen Chen, Minghui Qiu, W.~Bruce Croft, and Mohit Iyyer.
  2020.
\newblock \href {https://doi.org/10.1145/3397271.3401110} {Open-retrieval
  conversational question answering}.
\newblock In \emph{Proceedings of the 43rd International {ACM} {SIGIR}
  conference on research and development in Information Retrieval, {SIGIR}
  2020}, pages 539--548. {ACM}.

\bibitem[{Qu et~al.(2019)Qu, Yang, Qiu, Zhang, Chen, Croft, and
  Iyyer}]{DBLP:conf/cikm/QuYQZCCI19}
Chen Qu, Liu Yang, Minghui Qiu, Yongfeng Zhang, Cen Chen, W.~Bruce Croft, and
  Mohit Iyyer. 2019.
\newblock \href {https://doi.org/10.1145/3357384.3357905} {Attentive history
  selection for conversational question answering}.
\newblock In \emph{Proceedings of the 28th {ACM} International Conference on
  Information and Knowledge Management, {CIKM} 2019}, pages 1391--1400. {ACM}.

\bibitem[{Qu et~al.(2021)Qu, Ding, Liu, Liu, Ren, Zhao, Dong, Wu, and
  Wang}]{qu-etal-2021-rocketqa}
Yingqi Qu, Yuchen Ding, Jing Liu, Kai Liu, Ruiyang Ren, Wayne~Xin Zhao, Daxiang
  Dong, Hua Wu, and Haifeng Wang. 2021.
\newblock \href {https://doi.org/10.18653/v1/2021.naacl-main.466}
  {{R}ocket{QA}: An optimized training approach to dense passage retrieval for
  open-domain question answering}.
\newblock In \emph{Proceedings of the 2021 Conference of the North American
  Chapter of the Association for Computational Linguistics: Human Language
  Technologies}, pages 5835--5847, Online. Association for Computational
  Linguistics.

\bibitem[{Rajpurkar et~al.(2016)Rajpurkar, Zhang, Lopyrev, and Liang}]{squad}
Pranav Rajpurkar, Jian Zhang, Konstantin Lopyrev, and Percy Liang. 2016.
\newblock \href {https://doi.org/10.18653/v1/d16-1264} {Squad: 100, 000+
  questions for machine comprehension of text}.
\newblock In \emph{Proceedings of the 2016 Conference on Empirical Methods in
  Natural Language Processing, {EMNLP} 2016}, pages 2383--2392. The Association
  for Computational Linguistics.

\bibitem[{Raposo et~al.(2022)Raposo, Ribeiro, Martins, and
  Coheur}]{DBLP:conf/ecir/RaposoRMC22}
Gon{\c{c}}alo Raposo, Rui Ribeiro, Bruno Martins, and Lu{\'{\i}}sa Coheur.
  2022.
\newblock \href {https://doi.org/10.1007/978-3-030-99739-7\_23} {Question
  rewriting? assessing its importance for conversational question answering}.
\newblock In \emph{Advances in Information Retrieval - 44th European Conference
  on {IR} Research, {ECIR} 2022, Stavanger, Norway, April 10-14, 2022,
  Proceedings, Part {II}}, volume 13186 of \emph{Lecture Notes in Computer
  Science}, pages 199--206. Springer.

\bibitem[{Reddy et~al.(2019)Reddy, Chen, and
  Manning}]{DBLP:journals/tacl/ReddyCM19}
Siva Reddy, Danqi Chen, and Christopher~D. Manning. 2019.
\newblock \href {https://doi.org/10.1162/tacl\_a\_00266} {Coqa: {A}
  conversational question answering challenge}.
\newblock \emph{Trans. Assoc. Comput. Linguistics}, 7:249--266.

\bibitem[{Robertson et~al.(1994)Robertson, Walker, Jones, Hancock{-}Beaulieu,
  and Gatford}]{DBLP:conf/trec/RobertsonWJHG94}
Stephen~E. Robertson, Steve Walker, Susan Jones, Micheline Hancock{-}Beaulieu,
  and Mike Gatford. 1994.
\newblock \href {http://trec.nist.gov/pubs/trec3/papers/city.ps.gz} {Okapi at
  {TREC-3}}.
\newblock In \emph{Proceedings of The Third Text REtrieval Conference, {TREC}
  1994}, volume 500-225 of \emph{{NIST} Special Publication}, pages 109--126.
  National Institute of Standards and Technology {(NIST)}.

\bibitem[{Seo et~al.(2017)Seo, Kembhavi, Farhadi, and Hajishirzi}]{bidaf}
Min~Joon Seo, Aniruddha Kembhavi, Ali Farhadi, and Hannaneh Hajishirzi. 2017.
\newblock \href {https://openreview.net/forum?id=HJ0UKP9ge} {Bidirectional
  attention flow for machine comprehension}.
\newblock In \emph{5th International Conference on Learning Representations,
  {ICLR} 2017, Toulon, France, April 24-26, 2017, Conference Track
  Proceedings}.

\bibitem[{Seo et~al.(2019)Seo, Lee, Kwiatkowski, Parikh, Farhadi, and
  Hajishirzi}]{DBLP:conf/acl/SeoLKPFH19}
Min~Joon Seo, Jinhyuk Lee, Tom Kwiatkowski, Ankur~P. Parikh, Ali Farhadi, and
  Hannaneh Hajishirzi. 2019.
\newblock \href {https://doi.org/10.18653/v1/p19-1436} {Real-time open-domain
  question answering with dense-sparse phrase index}.
\newblock In \emph{Proceedings of the 57th Conference of the Association for
  Computational Linguistics, {ACL} 2019}, pages 4430--4441. Association for
  Computational Linguistics.

\bibitem[{Trischler et~al.(2017)Trischler, Wang, Yuan, Harris, Sordoni,
  Bachman, and Suleman}]{newsqa}
Adam Trischler, Tong Wang, Xingdi Yuan, Justin Harris, Alessandro Sordoni,
  Philip Bachman, and Kaheer Suleman. 2017.
\newblock \href {https://doi.org/10.18653/v1/w17-2623} {Newsqa: {A} machine
  comprehension dataset}.
\newblock In \emph{Proceedings of the 2nd Workshop on Representation Learning
  for NLP, Rep4NLP@ACL 2017}, pages 191--200. Association for Computational
  Linguistics.

\bibitem[{Vakulenko et~al.(2021)Vakulenko, Longpre, Tu, and
  Anantha}]{DBLP:conf/wsdm/VakulenkoLTA21}
Svitlana Vakulenko, Shayne Longpre, Zhucheng Tu, and Raviteja Anantha. 2021.
\newblock \href {https://doi.org/10.1145/3437963.3441748} {Question rewriting
  for conversational question answering}.
\newblock In \emph{{WSDM} '21, The Fourteenth {ACM} International Conference on
  Web Search and Data Mining, Virtual Event, Israel, March 8-12, 2021}, pages
  355--363. {ACM}.

\bibitem[{Vaswani et~al.(2017)Vaswani, Shazeer, Parmar, Uszkoreit, Jones,
  Gomez, Kaiser, and Polosukhin}]{DBLP:conf/nips/VaswaniSPUJGKP17}
Ashish Vaswani, Noam Shazeer, Niki Parmar, Jakob Uszkoreit, Llion Jones,
  Aidan~N. Gomez, Lukasz Kaiser, and Illia Polosukhin. 2017.
\newblock \href
  {https://proceedings.neurips.cc/paper/2017/hash/3f5ee243547dee91fbd053c1c4a845aa-Abstract.html}
  {Attention is all you need}.
\newblock In \emph{Advances in Neural Information Processing Systems 30: Annual
  Conference on Neural Information Processing Systems 2017}, pages 5998--6008.

\bibitem[{Wolf et~al.(2020)Wolf, Debut, Sanh, Chaumond, Delangue, Moi, Cistac,
  Rault, Louf, Funtowicz, Davison, Shleifer, von Platen, Ma, Jernite, Plu, Xu,
  Scao, Gugger, Drame, Lhoest, and Rush}]{DBLP:conf/emnlp/WolfDSCDMCRLFDS20}
Thomas Wolf, Lysandre Debut, Victor Sanh, Julien Chaumond, Clement Delangue,
  Anthony Moi, Pierric Cistac, Tim Rault, R{\'{e}}mi Louf, Morgan Funtowicz,
  Joe Davison, Sam Shleifer, Patrick von Platen, Clara Ma, Yacine Jernite,
  Julien Plu, Canwen Xu, Teven~Le Scao, Sylvain Gugger, Mariama Drame, Quentin
  Lhoest, and Alexander~M. Rush. 2020.
\newblock \href {https://doi.org/10.18653/v1/2020.emnlp-demos.6} {Transformers:
  State-of-the-art natural language processing}.
\newblock In \emph{Proceedings of the 2020 Conference on Empirical Methods in
  Natural Language Processing: System Demonstrations, {EMNLP} 2020 - Demos},
  pages 38--45. Association for Computational Linguistics.

\bibitem[{Wu et~al.(2022)Wu, Luan, Rashkin, Reitter, Hajishirzi, Ostendorf, and
  Tomar}]{https://doi.org/10.48550/arxiv.2112.08558}
Zeqiu Wu, Yi~Luan, Hannah Rashkin, David Reitter, Hannaneh Hajishirzi, Mari
  Ostendorf, and Gaurav~Singh Tomar. 2022.
\newblock \href {https://doi.org/10.48550/ARXIV.2112.08558} {C{O}{N}{Q}{R}{R}:
  Conversational query rewriting for retrieval with reinforcement learning}.
\newblock In \emph{Proceedings of the 2022 Conference on Empirical Methods in
  Natural Language Processing, {EMNLP} 2022}. Association for Computational
  Linguistics.

\bibitem[{Yu et~al.(2021)Yu, Liu, Xiong, Feng, and
  Liu}]{DBLP:conf/sigir/YuLXF021}
Shi Yu, Zhenghao Liu, Chenyan Xiong, Tao Feng, and Zhiyuan Liu. 2021.
\newblock \href {https://doi.org/10.1145/3404835.3462856} {Few-shot
  conversational dense retrieval}.
\newblock In \emph{{SIGIR} '21: The 44th International {ACM} {SIGIR} Conference
  on Research and Development in Information Retrieval}, pages 829--838. {ACM}.

\bibitem[{Zaib et~al.(2022)Zaib, Zhang, Sheng, Mahmood, and
  Zhang}]{DBLP:journals/kais/ZaibZSMZ22}
Munazza Zaib, Wei~Emma Zhang, Quan~Z. Sheng, Adnan Mahmood, and Yang Zhang.
  2022.
\newblock \href {https://doi.org/10.1007/s10115-022-01744-y} {Conversational
  question answering: a survey}.
\newblock \emph{Knowl. Inf. Syst.}, 64(12):3151--3195.

\end{thebibliography}
\bibliographystyle{acl_natbib}

%\appendix
%\section{Example Appendix}
%\label{sec:appendix}
%This is a section in the appendix.

\end{document}